\documentclass[11pt]{article}

\usepackage[preprint]{acl}

\usepackage{times}
\usepackage{latexsym}

\usepackage[T1]{fontenc}

\usepackage[utf8]{inputenc}

\usepackage{microtype}

\usepackage{inconsolata}

\usepackage{graphicx}
\usepackage{multirow}
\usepackage{sidecap}
\usepackage{float}
\usepackage{enumitem}

\title{Structurally Speaking: Motif-Oriented Graph Captioning through Bidirectional Graph-Text Translation}

\author{Hsiao-Ying Lu \and Dongyu Liu \and Kwan-Liu Ma \\
        \texttt{\{hyllu,dyuliu,klma\}@ucdavis.edu} \\
        Department of Computer Science, University of California, Davis}

\begin{document}
\maketitle
\begin{abstract}
    Graph captions should help readers understand graph structure, rather than simply translate adjacency matrices into long textual edge lists. A useful graph caption abstracts connectivity into recognizable motifs, such as hubs, paths, cycles, cliques, and bridges, because these motifs provide compact structural units that are easier to read, compare, and recover. In this paper, we study motif-oriented graph captioning as a bidirectional graph-text translation task, where captions must both preserve enough topology for graph recovery and express the graph through concise motif-level descriptions. We show that direct prompting of GPT-5.1 often produces graph-recoverable captions by enumerating node-to-node connections, but these captions are verbose and can contain inconsistent motif interpretations.
    To address this gap, we introduce \textbf{Structurally Speaking}, a lightweight structured prompting protocol that guides translation between explicit connectivity and motif-level abstraction. Experiments on a synthetic motif-based dataset show that structured prompting produces shorter and more motif-consistent captions while maintaining comparable graph recovery. These results suggest that explicit topology-to-motif reasoning guidance can make LLM-generated graph captions more interpretable without model fine-tuning.

\end{abstract}

\section{Introduction}

Graphs encode relationships among entities, but raw representations such as adjacency matrices are difficult to read and compare directly. Graph captions can make graph structure more accessible by translating connectivity into natural language. However, a useful caption should do more than list edges: it should abstract topology into recognizable motifs, such as hubs, paths, cycles, cliques, bridges, and tails. These motifs provide compact structural units that help readers understand how local connections compose into larger graph patterns.

This distinction matters because graph recoverability alone is not sufficient for evaluating graph-caption quality. A caption may encode all edges through enumeration while offering little interpretable structural abstraction. Such captions can be verbose, difficult to read, or inconsistent in their motif-level structural inference. We therefore argue that interpretable graph captioning should be evaluated along two dimensions: structural recoverability and motif-level compactness.

Motivated by this gap, we formulate \textit{bidirectional graph-caption translation} as a controlled diagnostic task. Given a graph adjacency matrix, the model generates a caption; given a caption, the model recovers the graph. This formulation allows us to probe whether LLMs can move between raw graph topology and motif-oriented natural language. Under direct prompting, GPT-5.1 often produces graph-recoverable captions, but these captions rely heavily on explicit node-to-node connection descriptions and sometimes contain inaccurate or self-contradictory motif interpretations. This suggests that direct prompting often falls back on edge enumeration rather than producing compact motif-level structural inference, even when graph recovery is successful.

To address this issue, we introduce \textbf{Structurally Speaking}, a lightweight structured prompting protocol for topology-to-motif abstraction. For graph-to-caption translation, the protocol guides the model to convert adjacency information into local neighborhoods, analyze motifs, and generate a compact caption. For caption-to-graph translation, it guides the model to parse motif descriptions, assign nodes, construct edges, and recover the adjacency matrix. This is not merely caption style transfer, because the model must infer motif structure from raw topology while preserving enough information for graph recovery.

We evaluate this formulation using a synthetic motif-based dataset and two cycle-consistency evaluations. Graph-Caption-Graph measures whether generated captions preserve enough structural information for graph recovery, while Caption-Graph-Caption measures whether motif-oriented descriptions remain accurate and compact after translation through graph structure. Experiments show that \textbf{Structurally Speaking} reduces verbosity and inconsistent motif inferences while maintaining comparable graph recovery. In summary, we contribute: (1) a bidirectional graph-captioning formulation that separates structural recoverability from motif-level abstraction; (2) a cycle-consistency evaluation that reveals recoverable captions can still be verbose and motif-inconsistent; and (3) a lightweight structured prompting protocol that improves caption compactness and motif consistency without fine-tuning.

\section{Related Work}

Recent work on LLMs and graphs spans three related directions. First, graph-to-text generation produces natural language from graph-structured inputs such as meaning representations, knowledge graphs, and scientific graphs~\cite{ribeiro-etal-2021-investigating}, with recent work evaluating LLMs for graph-to-text generation through planning and grounding-oriented tasks~\cite{he-etal-2025-evaluating}. While closely related, these works focus mainly on semantic graph structures and text fluency or factuality. We instead study motif-oriented captioning of raw topology, where captions should expose compact structural abstractions while remaining recoverable.

A second line of work instead optimizes how topology is serialized or represented for LLMs, including adjacency linearization~\cite{fatemi2023talk}, learned graph encodings for frozen LLMs~\cite{perozzi2024let}, and broader graph transformation strategies~\cite{10.1145/3786600}. A third line of work shifts from graph representation and generation to graph reasoning, learning, and query execution, including LLM-graph learning frameworks~\cite{10697304,11192225,you2025large}, graph-specialized prompting and structured interfaces~\cite{tang2024graphgpt,wang2024instructgraph,jiang2023structgpt,li2025graphotter}, empirical analyses of graph reasoning generalization~\cite{guo2023gpt4graph,zhang2024can}, and natural-language-to-graph-query systems~\cite{10.1145/3627673.3679713,hains:hal-04047470}.

\begin{figure*}
    \centering
    \includegraphics[width=\linewidth]{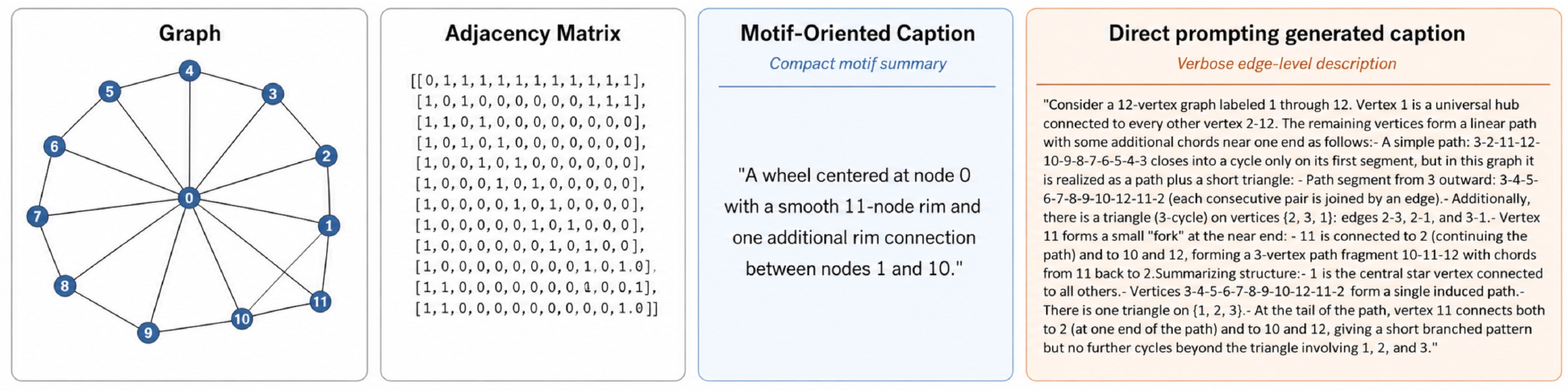}
    \caption{A wheel-motif graph with one added edge as structural variation. Its adjacency matrix is paired with a human-verified motif-oriented caption that compactly describes the central hub, rim structure, and additional rim connection. In contrast, the generated caption is graph-recoverable but verbose and sometimes self-contradictory, relying heavily on explicit node-to-node connection descriptions rather than compact motif-level abstraction.}
    \label{fig:IdealCaption}
\vspace{-0.1in}
\end{figure*}

\section{Problem Formulation}

Given a graph $G=(V,E)$, graph-to-caption translation aims to generate a natural-language caption $C$ that summarizes the topology of $G$. Given a caption $C$, caption-to-graph translation aims to reconstruct a graph $\hat{G}$. A high-quality caption should satisfy two criteria: \textit{graph recoverability} and \textit{motif abstraction}.

Specifically, we define a \textit{motif-oriented caption} as one that satisfies the following requirements:
\begin{itemize}[nosep]
    \item identifies dominant motifs, such as stars, paths, cycles, cliques, and wheels;
    \item describes node roles, such as hubs, rim nodes, bridge nodes, and leaves;
    \item describes deviations or perturbations, such as missing edges or added chords;
    \item avoids exhaustive edge enumeration unless specific connections are needed for graph recovery.
\end{itemize}

To support this task, we construct a synthetic motif-based graph dataset containing 220 undirected, unweighted graphs, each controlled to be within 30 nodes. Graphs are initialized from common motifs, including stars, cycles, paths, cliques, and wheels, and are then perturbed with random edge additions and removals to introduce structural variation. This process produces graphs with diverse topologies and varying levels of motif complexity. Illustrations are provided in \autoref{app:dataExamples}.

From these 220 graphs, we select 40 representative examples for caption annotation. These graphs are chosen to cover diverse motif families and perturbation levels. Each selected graph is paired with a human-verified motif-oriented caption, as shown in \autoref{fig:IdealCaption}. The captions are written in free-form natural language rather than templates to preserve linguistic diversity and allow natural rephrasings. This set of 40 graph-caption pairs serves as the unified test set for our experiments and analyses. It is intentionally curated for controlled diagnosis rather than benchmark-scale evaluation.

For LLM input, each graph is represented as an adjacency matrix serialized into text. This representation preserves explicit topology while remaining compatible with autoregressive language models.

\section{Direct Prompting Analysis}

We first examine GPT-5.1 under direct prompting to probe how a strong LLM approaches bidirectional graph-caption translation without task-specific guidance or fine-tuning. 

Our results (\autoref{tab:main_results}) show that direct prompting often preserves enough connectivity information for graph recovery. However, across captions, we observe three recurring failure modes: (1) edge enumeration instead of motif abstraction, (2) low caption compactness, and (3) inconsistent or inaccurate motif interpretation. \autoref{fig:IdealCaption} provides a representative example, annotated in \autoref{tab:direct_failure_modes}. These observations show that graph recoverability can be achieved through edge enumeration, and therefore does not by itself provide sufficient evidence of motif-level abstraction. This motivates a prompting strategy that separates connectivity extraction from motif abstraction before generating the final caption.

\begin{table}[t]
\centering
\small
\caption{Representative direct-prompting failure modes.}
\begin{tabular}{p{0.08\linewidth}p{0.82\linewidth}}
\hline
\textbf{Mode} & \textbf{Selected evidence in \autoref{fig:IdealCaption}} \\
\hline
(1) & ``3-2-11-12-10-9-8-7-6-5-4-3'', ``2-3'', ``2-1'', etc. \\
(2) & Direct caption is visibly much longer than the motif-oriented caption. \\
(3) & Alternates among path, star, triangle, and fork descriptions, obscuring the wheel motif. \\
\hline
\end{tabular}
\label{tab:direct_failure_modes}
\vspace{-0.1in}
\end{table}




\section{Structurally Speaking}
\label{sec:SS}

To better balance graph recovery and motif-level abstraction, we introduce \textbf{Structurally Speaking}, a structured chain-of-thought reasoning protocol for bidirectional graph-caption translation. This protocol is designed to enforce an intermediate representation between adjacency matrices and captions: neighbor lists expose explicit local connectivity, while motif analysis groups local edges into higher-level structural units. Instead of asking the model to directly produce captions or graphs in a single step, our protocol decomposes each translation direction into graph-specific reasoning stages before generating the final output.

The structured reasoning protocol follows the prompt templates below. The exact prompts used are provided in \autoref{app:prompts}. For graph-to-caption translation, we use:



{\vspace{0.1in}
\small
\noindent
\begin{tabular}{@{}p{0.12\linewidth}p{0.82\linewidth}@{}}
\textit{Step 1:} & \textit{Convert adjacency matrix to neighbor list (0-indexed).} \\
\textit{Step 2:} & Analyze structure (i.e., what motifs are in this pattern). \\
\textit{Step 3:} & Generate final caption. \\
\end{tabular}
}

For caption-to-graph translation, we use:

{\vspace{0.1in}
\small
\noindent
\begin{tabular}{@{}p{0.12\linewidth}p{0.82\linewidth}@{}}
\textit{Step 1:} & \textit{Parse structural descriptions (i.e., identify the number of nodes and motifs).} \\
\textit{Step 2:} & \textit{Assign indices and layout (i.e., assign nodes to different motifs).} \\
\textit{Step 3:} & \textit{Create edge list (i.e., analyze the edge required to construct the motifs).} \\
\textit{Step 4:} & \textit{Build neighbor list for each node.} \\
\textit{Step 5:} & \textit{Convert neighbor list to adj matrix.} \\
\end{tabular}
}

\begin{table*}[t]
\centering
\small
\caption{
Cycle-consistency evaluation using GPT-5.1 for all prompting methods. 
Graph-Caption-Graph evaluates graph recovery using edge precision, recall, and F1. 
Caption-Graph-Caption evaluates caption reconstruction using ROUGE-1 precision and recall, and caption compactness using average generated caption length (in characters). 
}
\begin{tabular}{lccc|ccc}
\hline
\multirow{2}{*}{\textbf{Prompting Method}} 
& \multicolumn{3}{c|}{\textbf{Graph-Caption-Graph}} 
& \multicolumn{3}{c}{\textbf{Caption-Graph-Caption}} \\
\cline{2-7}
& Precision & Recall & F1 
& ROUGE-1 Precision & ROUGE-1 Recall & Avg. Length \\
\hline
Direct Prompting & 1.0 & 1.0 & 1.0 & 0.07636 & 0.50933 & 1222.1 \\
Zero-shot Structured & 1.0 & 0.95751 & 0.97172 & 0.22378 & 0.55846 & 313.225 \\
Few-shot Structured & 0.99688 & 0.99688 & 0.99667 & 0.45166 & 0.53706 & 136.175 \\
\hline
\end{tabular}
\label{tab:main_results}
\vspace{-0.1in}
\end{table*}

The graph-to-caption template moves from connectivity extraction to motif-level abstraction, while the caption-to-graph template maps motif-oriented language back to explicit topology. We compare three prompting settings. Direct prompting uses only the task instruction. Notably, the direct prompt already asks the model to describe graph motifs, as shown in \autoref{app:prompts}. Zero-shot structured prompting uses the \textbf{Structurally Speaking} templates without labeled examples, testing the reasoning scaffold alone. Few-shot structured prompting uses the same templates with example graph-caption pairs with full human-verified intermediate reasoning paths and motif-oriented captions, serving as a demonstration-informed upper bound for this task; these examples are disjoint from the 40-example test set. All methods use the same underlying GPT-5.1 model and fixed decoding settings~\cite{singh2025openai}.



\section{Evaluation}

\paragraph{Cycle-Consistency Evaluation.}
We evaluate two cycle-consistency settings. In \textit{Graph-Caption-Graph}, the model generates a caption from a graph and then reconstructs a graph from that caption. This measures graph recovery, i.e., whether the caption preserves enough information to recover the original topology. Given the original edge set $E$ and reconstructed edge set $\hat{E}$, we compute edge precision $P=\frac{|E\cap\hat{E}|}{|\hat{E}|}$, recall $R=\frac{|E\cap\hat{E}|}{|E|}$, and $F1=\frac{2PR}{P+R}$.

\vspace{0.1in}
In \textit{Caption-Graph-Caption}, the model reconstructs a graph from a reference caption and then generates a new caption from the reconstructed graph. This measures whether motif-oriented descriptions remain accurate and compact after translation through graph structure. We treat the human-written motif-oriented caption as the reference and the model-generated caption as the hypothesis. We use ROUGE~\cite{lin-2004-rouge} as a lightweight proxy for lexical overlap with human-verified motif captions, interpreted together with caption length (in characters) and qualitative inspection.  We use the \texttt{rouge-score} implementation with stemming enabled. ROUGE-1 recall measures how much motif-relevant content is covered, while ROUGE-1 precision reflects how much additional wording is introduced. Thus, lower precision with longer captions may indicate verbose adjacency-oriented details rather than missing motif content alone. Together, these cycles evaluate both structural faithfulness and motif-level abstraction quality.

\paragraph{Structured Reasoning Analysis.}
As introduced in \autoref{sec:SS}, we evaluate three prompting schemes using GPT-5.1. \autoref{tab:main_results} shows three main trends.

First, \textit{direct prompting} achieves perfect edge precision, recall, and F1 in Graph-Caption-Graph, but produces the longest captions and lowest ROUGE-1 precision in Caption-Graph-Caption. This indicates that recovery is driven mainly by edge enumeration rather than motif abstraction. Its moderate ROUGE-1 recall may partly reflect this verbosity: the generated captions mention several motif-relevant terms, increasing lexical coverage, but these terms are sometimes embedded in inaccurate or self-contradictory motif interpretations.

Second, \textit{zero-shot structured prompting} greatly reduces caption length and improves ROUGE-1 precision and recall over direct prompting, suggesting that the \textbf{Structurally Speaking} scaffold helps elicit more focused topology-to-motif abstraction without any labeled examples. Its small drop in Graph-Caption-Graph recall indicates that more compact captions may omit some edge details.

Third, \textit{few-shot structured prompting} maintains near-perfect graph recovery while producing the shortest captions and highest ROUGE-1 precision. This suggests that demonstrations help align the desired output convention with motif-level abstraction, enabling compact captions without sacrificing recoverability. Qualitative examples in \autoref{fig:IdealCaption} and \autoref{app:additional_qual} further show that structured prompting results in shorter captions that actually correspond to better motif abstraction, not just shorter text.

We therefore interpret the few-shot setting as a demonstration-informed upper-bound condition, while the zero-shot setting more directly tests the benefit of our structured reasoning scaffold. Overall, these results suggest that structured prompting helps balance graph recoverability with compact and accurate motif-oriented captioning. Additional detailed results are provided in \autoref{app:additional_quant}.

\section{Conclusion}

We studied graph captioning as a diagnostic probe of whether LLMs can infer motif-level structure from raw graph topology. Through bidirectional graph-caption translation, we evaluated both graph recovery and motif-oriented caption compactness. Our cycle-consistency evaluation shows that direct prompting can preserve topology through verbose edge enumeration, but does not necessarily reflect interpretable motif-level abstraction. 
\textbf{Structurally Speaking} reduces this mismatch by guiding topology-to-motif reasoning, eliciting more accurate and consistent motif-level structural inference while maintaining comparable graph recovery.


\section*{Limitations}
This work is a controlled diagnostic study rather than a comprehensive benchmark. Our experiments use a synthetic motif-based dataset and a single LLM, so the results may not generalize to all graph families, larger graphs, or other models. The captioned evaluation set is also small because motif-oriented captions require human verification. In addition, ROUGE-1 and caption length provide only approximate signals of motif-oriented caption quality; we therefore interpret them together with graph recovery metrics and qualitative examples. Future work should expand the dataset, evaluate additional models, and incorporate human judgments of caption usefulness and abstraction level.

\section*{Ethical Considerations}

This work uses synthetic graph structures and human-written motif-oriented captions, and does not involve personal, sensitive, or human-subject data. As such, we do not identify direct ethical risks from the dataset itself. The main consideration is a general risk of LLM-based graph-language systems: generated captions may appear fluent and plausible while omitting, distorting, or over-specifying structural information. We address this risk in our study by evaluating captions through graph recovery, caption reconstruction, caption length, and qualitative inspection rather than relying on fluency alone. In practical applications, generated graph captions should be verified against the underlying graph before being used for analysis or decision-making.


\bibliography{00_ref}

@inproceedings{he-etal-2025-evaluating,
    title = "Evaluating and Improving Graph to Text Generation with Large Language Models",
    author = "He, Jie  and
      Yang, Yijun  and
      Long, Wanqiu  and
      Xiong, Deyi  and
      Gutierrez Basulto, Victor  and
      Pan, Jeff Z.",
    editor = "Chiruzzo, Luis  and
      Ritter, Alan  and
      Wang, Lu",
    booktitle = "Proceedings of the 2025 Conference of the Nations of the Americas Chapter of the Association for Computational Linguistics: Human Language Technologies (Volume 1: Long Papers)",
    month = apr,
    year = "2025",
    address = "Albuquerque, New Mexico",
    publisher = "Association for Computational Linguistics",
    url = "https://aclanthology.org/2025.naacl-long.513/",
    doi = "10.18653/v1/2025.naacl-long.513",
    pages = "10219--10244"
}

@inproceedings{ribeiro-etal-2021-investigating,
    title = "Investigating Pretrained Language Models for Graph-to-Text Generation",
    author = "Ribeiro, Leonardo F. R.  and
      Schmitt, Martin  and
      Sch{\"u}tze, Hinrich  and
      Gurevych, Iryna",
    editor = "Budzianowski, Pawe{\l}  and
      Gasic, Milica  and
      Minlie, Huang  and
      Vuli{\'c}, Ivan  and
      Zhang, Chen  and
      Casanueva, I{\~n}igo  and
      Wen, Tsung-Hsien  and
      Su, Pei-Hao  and
      Ultes, Stefan",
    booktitle = "Proceedings of the 3rd Workshop on Natural Language Processing for Conversational AI",
    month = nov,
    year = "2021",
    address = "Online",
    publisher = "Association for Computational Linguistics",
    url = "https://aclanthology.org/2021.nlp4convai-1.20/",
    doi = "10.18653/v1/2021.nlp4convai-1.20",
    pages = "211--227"
}

@article{singh2025openai,
  title={Openai gpt-5 system card},
  author={Singh, Aaditya and Fry, Adam and Perelman, Adam and Tart, Adam and Ganesh, Adi and El-Kishky, Ahmed and McLaughlin, Aidan and Low, Aiden and Ostrow, AJ and Ananthram, Akhila and others},
  journal={arXiv preprint arXiv:2601.03267},
  year={2025}
}

@inproceedings{lin-2004-rouge,
    title = "{ROUGE}: A Package for Automatic Evaluation of Summaries",
    author = "Lin, Chin-Yew",
    booktitle = "Text Summarization Branches Out",
    month = jul,
    year = "2004",
    address = "Barcelona, Spain",
    publisher = "Association for Computational Linguistics",
    url = "https://aclanthology.org/W04-1013/",
    pages = "74--81"
}

@inproceedings{hains:hal-04047470,
  TITLE = {{From natural language to graph queries}},
  AUTHOR = {Hains, Gaetan and Khmelevsky, Youry and Tachon, Thibaut},
  URL = {https://hal.science/hal-04047470},
  BOOKTITLE = {{2019 IEEE Canadian Conference of Electrical and Computer Engineering (CCECE)}},
  ADDRESS = {Edmonton, Canada},
  PUBLISHER = {{IEEE}},
  PAGES = {1-4},
  YEAR = {2019},
  MONTH = May,
  DOI = {10.1109/CCECE.2019.8861892},
  HAL_ID = {hal-04047470},
  HAL_VERSION = {v1},
}

@inproceedings{10.1145/3627673.3679713,
author = {Liang, Yuanyuan and Tan, Keren and Xie, Tingyu and Tao, Wenbiao and Wang, Siyuan and Lan, Yunshi and Qian, Weining},
title = {Aligning Large Language Models to a Domain-specific Graph Database for NL2GQL},
year = {2024},
isbn = {9798400704369},
publisher = {Association for Computing Machinery},
address = {New York, NY, USA},
url = {https://doi.org/10.1145/3627673.3679713},
doi = {10.1145/3627673.3679713},
booktitle = {Proceedings of the 33rd ACM International Conference on Information and Knowledge Management},
pages = {1367–1377},
numpages = {11},
location = {Boise, ID, USA},
series = {CIKM '24}
}

@ARTICLE{11192225,
  author={Shang, Wenbo and Huang, Xin},
  journal={IEEE Transactions on Knowledge and Data Engineering}, 
  title={A Survey of Large Language Models on Generative Graph Analytics: Query, Learning, and Applications}, 
  year={2025},
  volume={37},
  number={12},
  pages={6799-6819},
  doi={10.1109/TKDE.2025.3609877}}

@article{10.1145/3786600,
author = {Yu, Shuo and Wang, Yingbo and Li, Ruolin and Liu, Guchun and Shen, Yanming and Ji, Shaoxiong and Li, Bowen and Han, Fengling and Zhang, Xiuzhen and Xia, Feng},
title = {Graph2text or Graph2token: A Perspective of Large Language Models for Graph Learning},
year = {2026},
issue_date = {March 2026},
publisher = {Association for Computing Machinery},
address = {New York, NY, USA},
volume = {44},
number = {3},
issn = {1046-8188},
url = {https://doi.org/10.1145/3786600},
doi = {10.1145/3786600},
journal = {ACM Trans. Inf. Syst.},
month = feb,
articleno = {57},
numpages = {49}
}

@inproceedings{tang2024graphgpt,
  title={Graphgpt: Graph instruction tuning for large language models},
  author={Tang, Jiabin and Yang, Yuhao and Wei, Wei and Shi, Lei and Su, Lixin and Cheng, Suqi and Yin, Dawei and Huang, Chao},
  booktitle={Proceedings of the 47th International ACM SIGIR Conference on Research and Development in Information Retrieval},
  pages={491--500},
  year={2024}
}

@inproceedings{wang2024instructgraph,
  title={Instructgraph: Boosting large language models via graph-centric instruction tuning and preference alignment},
  author={Wang, Jianing and Wu, Junda and Hou, Yupeng and Liu, Yao and Gao, Ming and McAuley, Julian},
  booktitle={Findings of the Association for Computational Linguistics: ACL 2024},
  pages={13492--13510},
  year={2024}
}

@ARTICLE{10697304,
  author={Jin, Bowen and Liu, Gang and Han, Chi and Jiang, Meng and Ji, Heng and Han, Jiawei},
  journal={IEEE Transactions on Knowledge and Data Engineering}, 
  title={Large Language Models on Graphs: A Comprehensive Survey}, 
  year={2024},
  volume={36},
  number={12},
  pages={8622-8642},
  doi={10.1109/TKDE.2024.3469578}}

@article{you2025large,
  title={Large language models meet graph neural networks: a perspective of graph mining},
  author={You, Yuxin and Liu, Zhen and Wen, Xiangchao and Zhang, Yongtao and Ai, Wei},
  journal={Mathematics},
  volume={13},
  number={7},
  pages={1147},
  year={2025},
  publisher={MDPI}
}

@inproceedings{li2025graphotter,
  title={Graphotter: Evolving llm-based graph reasoning for complex table question answering},
  author={Li, Qianlong and Huang, Chen and Li, Shuai and Xiang, Yuanxin and Xiong, Deng and Lei, Wenqiang},
  booktitle={Proceedings of the 31st International Conference on Computational Linguistics},
  pages={5486--5506},
  year={2025}
}

@inproceedings{zhang2024can,
  title={Can LLM graph reasoning generalize beyond pattern memorization?},
  author={Zhang, Yizhuo and Wang, Heng and Feng, Shangbin and Tan, Zhaoxuan and Han, Xiaochuang and He, Tianxing and Tsvetkov, Yulia},
  booktitle={Findings of the Association for Computational Linguistics: EMNLP 2024},
  pages={2289--2305},
  year={2024}
}

@article{fatemi2023talk,
  title={Talk like a graph: Encoding graphs for large language models},
  author={Fatemi, Bahare and Halcrow, Jonathan and Perozzi, Bryan},
  journal={arXiv preprint arXiv:2310.04560},
  year={2023}
}

@article{perozzi2024let,
  title={Let your graph do the talking: Encoding structured data for llms},
  author={Perozzi, Bryan and Fatemi, Bahare and Zelle, Dustin and Tsitsulin, Anton and Kazemi, Mehran and Al-Rfou, Rami and Halcrow, Jonathan},
  journal={arXiv preprint arXiv:2402.05862},
  year={2024}
}

@article{guo2023gpt4graph,
  title={Gpt4graph: Can large language models understand graph structured data? an empirical evaluation and benchmarking},
  author={Guo, Jiayan and Du, Lun and Liu, Hengyu and Zhou, Mengyu and He, Xinyi and Han, Shi},
  journal={arXiv preprint arXiv:2305.15066},
  year={2023}
}

@inproceedings{jiang2023structgpt,
  title={Structgpt: A general framework for large language model to reason over structured data},
  author={Jiang, Jinhao and Zhou, Kun and Dong, Zican and Ye, Keming and Zhao, Wayne Xin and Wen, Ji-Rong},
  booktitle={Proceedings of the 2023 Conference on Empirical Methods in Natural Language Processing},
  pages={9237--9251},
  year={2023}
}

\appendix

\section{Motif-Based Graphs}
\label{app:dataExamples}
\begin{figure}[!h]
    \centering
    \includegraphics[width=0.9\linewidth]{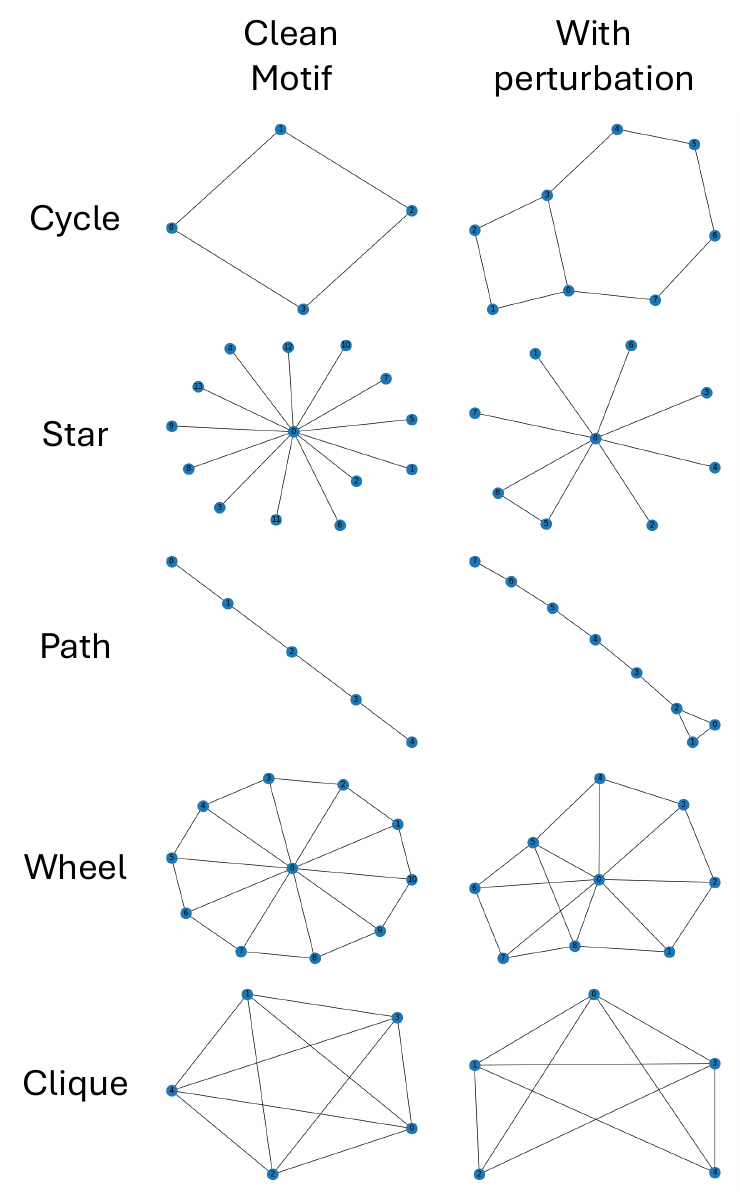}
    \caption{Examples from our synthetic motif-based graph dataset. Each row corresponds to a base motif family: cycle, star, path, wheel, and clique. The left column shows clean motif examples, while the right column shows examples from the same motif family with a single edge addition or removal perturbation. The two columns illustrate clean and perturbed cases but do not necessarily depict paired versions of the same graph.}
    \label{fig:motif_data_example}
\end{figure}

\autoref{fig:motif_data_example} shows examples from our synthetic motif-based graph dataset.

\begin{figure*}
    \centering
    \includegraphics[width=\linewidth]{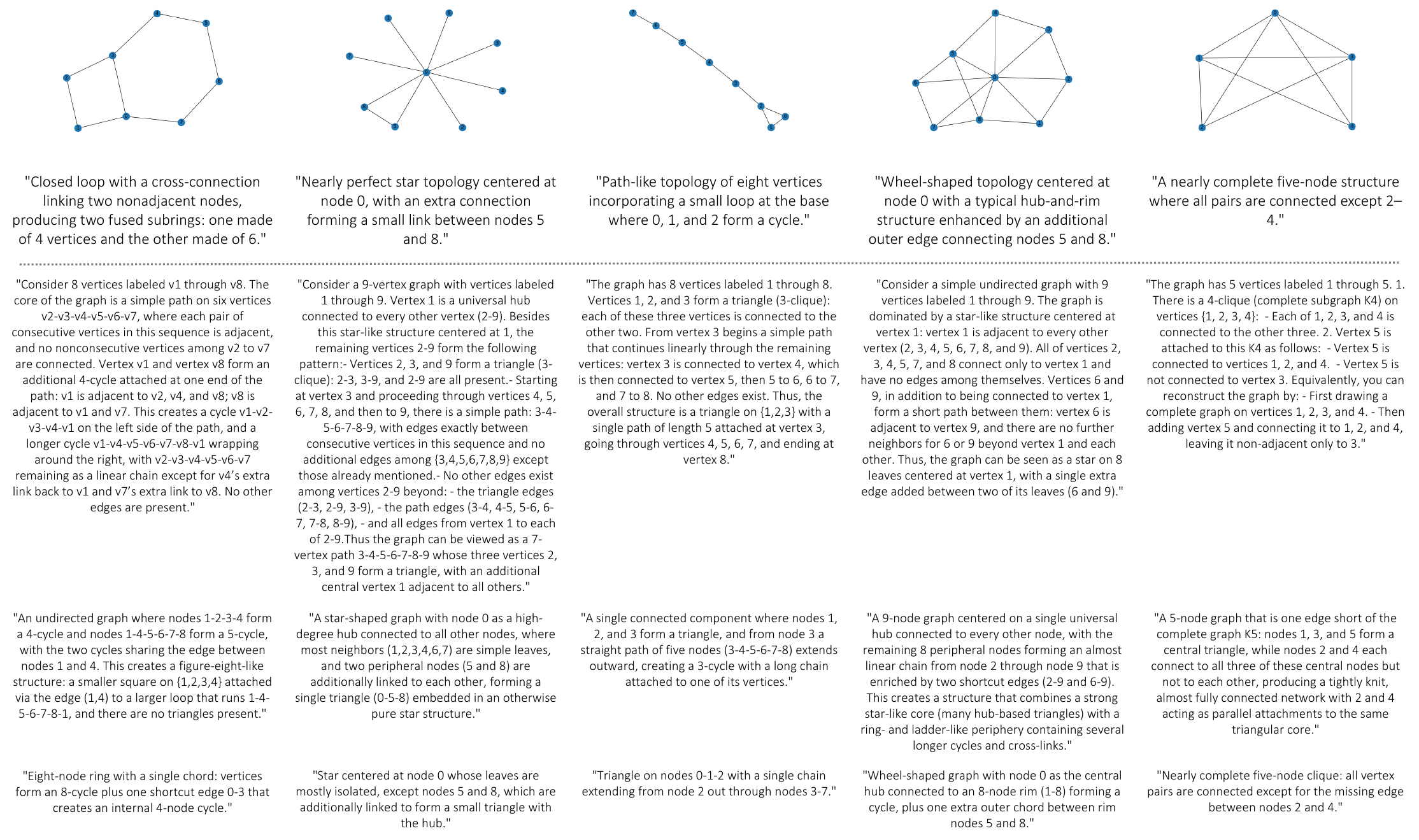}
    \caption{Examples of generated captions across prompting settings. The top row shows perturbed motif graphs with human-verified motif-oriented captions, followed by captions generated by direct prompting, zero-shot structured prompting, and few-shot structured prompting. Structured prompting reduces verbosity and self-contradictory descriptions, while better focusing on motif-oriented graph patterns.}
    \label{fig:additionalQual}
\end{figure*}

\section{Prompting Schemes}
\label{app:prompts}

For the three prompting schemes evaluated in this paper, we use the following exact instructions to prompt GPT-5.1.

\begin{itemize}
    \item Direct Prompting: \\
        {
            \small
            \noindent
            \begin{tabular}{@{}p{0.12\linewidth}p{0.82\linewidth}@{}}
            \textit{Caption to Graph:} & \textit{You are to generate a graph pattern in its adjacency matrix form that matches the given caption. Please write the adjacency matrix in a clear format, starting with 'Adjacency Matrix:' followed by the matrix itself, expressed as list of lists in string form.} \\
            \textit{Graph to Caption:} & \textit{You are to generate graph pattern caption describing the graph motifs from the given adjacency matrix. Start the output with 'Caption:' and then write the caption.} 
            \end{tabular}
        }
    \item Zero-shot structured prompting: \\
        {
            \small
            \noindent
            \begin{tabular}{@{}p{0.12\linewidth}p{0.82\linewidth}@{}}
            \textit{Caption to Graph:} & \textit{You are to generate a graph pattern that matches the above given caption. Please reason through these steps: step 1: Parse structural descriptions (i.e., identify the number of nodes and motifs), step 2: Assign indices and layout (i.e., assign nodes to different motifs), step 3: Create edge list (i.e., analyze the edge required to construct the motifs), step 4: Build neighbor list for each node, step 5: convert neighbor list to adj matrix. Please output the reasoning steps for a given graph pattern caption and start the output for the last step with 'Adjacency Matrix: ' followed by the matrix itself, expressed as list of lists in string form.} \\
            \textit{Graph to Caption:} & \textit{You are to generate graph pattern captions. Please reason through these steps: step 1: convert adjacency to produce neighbor lists, step 2: analyze structure (i.e., what motifs are in this pattern), step 3: generate final caption. Please output the reasoning steps for a given adjacency matrix and start the output for the last step with 'Caption: '.} 
            \end{tabular}
        }
    \item Few-shot structured prompting: \\
        {
            \small
            \noindent
            \begin{tabular}{@{}p{0.12\linewidth}p{0.82\linewidth}@{}}
            \textit{Caption to Graph:} & \textit{You are to generate a graph pattern that matches the above given caption. As you can observe in assistant examples: step 1: Parse structural descriptions (i.e., identify the number of nodes and motifs), step 2: Assign indices and layout (i.e., assign nodes to different motifs), step 3: Create edge list (i.e., analyze the edge required to construct the motifs), step 4: Build neighbor list for each node, step 5: convert neighbor list to adj matrix. Please learn from those assistant examples and output these reasoning steps and the final adj matrix for a given graph pattern caption.} \\
            \textit{Graph to Caption:} & \textit{You are to generate graph pattern captions. As you can observe in assistant examples: step 1: convert adjacency to produce neighbor lists, step 2: analyze structure (i.e., what motifs are in this pattern), step 3: generate final caption. Please learn from those assistant examples and output the reasoning through these steps reaching the final caption for a given adjacency matrix.} 
            \end{tabular}
        }
\end{itemize}

\begin{table*}[!ht]
\centering
\scriptsize
\caption{
Cycle-consistency evaluation on clean and perturbed motif graphs using GPT-5.1, where
perturbed graphs contain random edge additions or removals from the base motif structure.
}
\begin{tabular}{llccc|ccc}
\hline
\multirow{2}{*}{\textbf{Graph Type}} 
& \multirow{2}{*}{\textbf{Prompting Method}} 
& \multicolumn{3}{c|}{\textbf{Graph-Caption-Graph}} 
& \multicolumn{3}{c}{\textbf{Caption-Graph-Caption}} \\
\cline{3-8}
& & Precision & Recall & F1 
& ROUGE-1 Precision & ROUGE-1 Recall & Avg. Length \\
\hline
\multirow{3}{*}{Clean}
& Direct Prompting & 1.0 & 1.0 & 1.0 & 0.100823 & 0.500799 & 665.5 \\
& Zero-shot Structured & 1.0 & 1.0 & 1.0 & 0.266728 & 0.571888 & 200.7 \\
& Few-shot Structured & 1.0 & 1.0 & 1.0 & 0.575181 & 0.631384 & 94.5 \\
\hline
\multirow{3}{*}{Perturbed}
& Direct Prompting & 1.0 & 1.0 & 1.0 & 0.068201 & 0.512170 & 1407.633 \\
& Zero-shot Structured & 1.0 & 0.943351 & 0.962294 & 0.209464 & 0.553990 & 350.733 \\
& Few-shot Structured & 0.995833 & 0.995833 & 0.995556 & 0.410488 & 0.505617 & 150.067 \\
\hline
\end{tabular}
\label{tab:clean_perturbed_results}
\end{table*}

\begin{table*}[!ht]
\centering
\scriptsize
\caption{
Cycle-consistency evaluation by base motif family using GPT-5.1. 
}
\begin{tabular}{llccc|ccc}
\hline
\multirow{2}{*}{\textbf{Motif}} 
& \multirow{2}{*}{\textbf{Prompting Method}} 
& \multicolumn{3}{c|}{\textbf{Graph-Caption-Graph}} 
& \multicolumn{3}{c}{\textbf{Caption-Graph-Caption}} \\
\cline{3-8}
& & Precision & Recall & F1 
& ROUGE-1 Precision & ROUGE-1 Recall & Avg. Length \\
\hline
\multirow{3}{*}{Cycle}
& Direct Prompting & 1.0 & 1.0 & 1.0 & 0.070145 & 0.484059 & 1078.444 \\
& Zero-shot Structured & 1.0 & 1.0 & 1.0 & 0.243284 & 0.557181 & 357.111 \\
& Few-shot Structured & 0.986111 & 1.0 & 0.992593 & 0.526199 & 0.538382 & 121.889 \\
\hline
\multirow{3}{*}{Star}
& Direct Prompting & 1.0 & 1.0 & 1.0 & 0.067775 & 0.560572 & 1395.375 \\
& Zero-shot Structured & 1.0 & 0.848930 & 0.895803 & 0.263899 & 0.566327 & 292.375 \\
& Few-shot Structured & 1.0 & 0.984375 & 0.991667 & 0.409470 & 0.532376 & 154.375 \\
\hline
\multirow{3}{*}{Path}
& Direct Prompting & 1.0 & 1.0 & 1.0 & 0.083385 & 0.424597 & 781.2 \\
& Zero-shot Structured & 1.0 & 1.0 & 1.0 & 0.292959 & 0.595345 & 228.2 \\
& Few-shot Structured & 1.0 & 1.0 & 1.0 & 0.514075 & 0.560210 & 115.6 \\
\hline
\multirow{3}{*}{Wheel}
& Direct Prompting & 1.0 & 1.0 & 1.0 & 0.062481 & 0.507626 & 1780.545 \\
& Zero-shot Structured & 1.0 & 0.955372 & 0.972944 & 0.165656 & 0.549543 & 369.182 \\
& Few-shot Structured & 1.0 & 1.0 & 1.0 & 0.400842 & 0.542979 & 148.636 \\
\hline
\multirow{3}{*}{Clique}
& Direct Prompting & 1.0 & 1.0 & 1.0 & 0.110122 & 0.535666 & 664.167 \\
& Zero-shot Structured & 1.0 & 1.0 & 1.0 & 0.194631 & 0.545271 & 256.333 \\
& Few-shot Structured & 1.0 & 1.0 & 1.0 & 0.444836 & 0.480310 & 126.5 \\
\hline
\end{tabular}
\label{tab:motif_wise_results}
\end{table*}

\section{Additional Qualitative Examples}
\label{app:additional_qual}

\autoref{fig:additionalQual} presents additional examples comparing generated captions across prompting settings. These examples illustrate how the captions differ in motif abstraction, structural accuracy, specificity, and length. Overall, these supplementary qualitative results further support our claim that structured prompting reduces verbosity and self-contradictory descriptions while producing captions that better capture motif-oriented graph patterns.

\section{Additional Quantitative Analyses}
\label{app:additional_quant}

We further analyze model performance across graph perturbation settings and motif families. These analyses use the same evaluation metrics as in the main paper: Graph-Caption-Graph F1 for graph recovery, ROUGE-1 precision and recall for Caption-Graph-Caption caption reconstruction, and average generated caption length for caption compactness.

As shown in \autoref{tab:clean_perturbed_results}, all prompting schemes perform better on clean motif graphs than on perturbed graphs. This is expected because edge additions and removals introduce structural variation beyond the base motif patterns. To avoid inflating the aggregate results with many easy clean examples, our test set includes only a small number of clean graphs, with two clean examples per motif family. As a result, the overall results in \autoref{tab:main_results} more closely reflect performance on perturbed graphs, which better approximate the structural irregularities found in natural graph patterns.

\autoref{tab:motif_wise_results} further shows that Wheel patterns are among the most challenging motif families across prompting schemes. This is also expected because a wheel combines a central hub with a rim structure, requiring the model to recognize both star-like and cycle-like organization and their composition. Nevertheless, the Wheel-specific results are broadly consistent with the overall trends in \autoref{tab:main_results}: structured prompting improves caption compactness and motif-oriented reconstruction while maintaining comparable graph recovery. This suggests that \textbf{Structurally Speaking} helps guide the model through motif composition even for more structurally complex patterns such as wheels.

\end{document}